\documentclass[10pt,twocolumn,letterpaper]{article}

\usepackage{iccv}              

\definecolor{iccvblue}{rgb}{0.21,0.49,0.74}
\usepackage[pagebackref,breaklinks,colorlinks,allcolors=iccvblue]{hyperref}

\usepackage{tikz}
\usetikzlibrary{positioning,calc}
\usepackage{graphicx}
\usepackage{enumitem}
\usepackage{subcaption}
\usepackage[inkscapelatex=true]{svg}

\def\paperID{6} 
\def\confName{ICCV}
\def\confYear{2025}

\title{Interpretable Open-Vocabulary Referring Object Detection with\\ Reverse Contrast Attention}

\author{
Drandreb Earl O. Juanico\textsuperscript{1}, 
Rowel O. Atienza\textsuperscript{1,2}, 
Jeffrey Kenneth Go\textsuperscript{3}\\
\textsuperscript{1}AI Graduate Program, University of the Philippines, Diliman, Quezon City\\
\textsuperscript{2}EEEI, University of the Philippines, Diliman, Quezon City\\
\textsuperscript{3}Samsung R\&D Institute Philippines\\
{\tt\small dojuanico@up.edu.ph, earl.juanico@gmail.com, rowel@eee.upd.edu.ph, jeff.go@samsung.com}
}

\begin{document}
\maketitle
\begin{abstract}
We propose Reverse Contrast Attention (RCA), a plug-in method that enhances object localization in vision-language transformers without retraining. RCA reweights final-layer attention by suppressing extremes and amplifying mid-level activations to let semantically relevant but subdued tokens guide predictions. We evaluate it on Open Vocabulary Referring Object Detection (OV-RefOD), introducing FitAP—a confidence-free average precision metric based on IoU and box area. RCA improves FitAP in 11 out of 15 open-source VLMs, with gains up to $+26.6\%$. Effectiveness aligns with attention sharpness and fusion timing; while late-fusion models benefit consistently, models like DeepSeek-VL2 also improve, pointing to capacity and disentanglement as key factors. RCA offers both interpretability and performance gains for multimodal transformers.
\end{abstract}

\section{Introduction}
\label{sec:intro}

Vision-language transformers are widely considered as effective computational models for studying how natural language reasoning interfaces with visual perception. These models align and integrate information across image and text with multi-modal attention mechanisms. However, the interpretability of outputs by a vision-language model (VLM), especially in explaining how cross-modal attention pathways selectively propagate visual features in response to linguistic cues, remains a critical and ongoing research challenge. Underscoring this challenge is a debate sparked by two highly influential work. Jain and Wallace's ``\emph{Attention is Not Explanation}"~\cite{jain2019attention} demonstrated how significant modifications to the attention weights did not change the model output or responses. This conclusion undermines previous findings supporting the argument that attention reliably represents model reasoning. On the contrary, in their equally influential work, ``\emph{Attention is Not Not Explanation}", Wiegreffe and Pinter~\cite{wiegreffe2019attention} argued that in particular methodically crafted, constrained scenarios, attention may provide insightful interpretation.

The central debate in both influential papers concerns whether or not attention weights uniquely explain a VLM output. As Jain and Wallace demonstrated, multiple distinct attention distributions can often yield the same model response, indicating that no single configuration of attention weights acts as a definitive trace of the model's internal reasoning. In contrast, Wiegreffe and Pinter contended that attention can still serve as a useful interpretive tool, provided that modifications to the attention distributions are made deliberately, under constraints that preserve output fidelity and align with known model behavior. Taken together, these perspectives suggest that attention flexibility (the capacity to vary without altering predictions) can be productively used for interpretability. This functional plasticity of attention may thus be viewed not as a limitation but as a resource for probing the model's internal decision structure.

In this study, we propose to improve VLM performance in a computer vision task without additional training or fine-tuning by leveraging attention functional plasticity to directly manipulate the attention weights used to compute logits during inference. This manipulation preserves the gains from extensive pretraining on large-scale data sets while enabling inference-time adaptation to task-specific objectives. Building on previous eXCV studies focused on mapping and visualizing VLM attention, our method capitalizes on this attention-derived guidance for performance improvement, thus linking interpretability to functional enhancement. 

We establish this improvement by defining the task and the probing mechanism to achieve it in such a task. We then discuss how the manner of attention manipulation relates to VLM explainability. In particular, we draw parallels to the relevance propagation idea to interpret bi-modal transformers~\cite{chefer2021generic} and the tracer scoring approach with deep Taylor decomposition~\cite{chefer2021transformer}, which recognize the potential of attention traces not just for explanation, but also for actionable model intervention during inference.


\subsection{Related Work}
Reverse Contrast Attention (RCA) generalizes the focus shifting principle introduced by Chen \etal~\cite{chen2018reverse}, in which confident predictions are erased to help recover missed object regions in a top-down manner. Similarly, Huang \etal~\cite{huang2017semantic} employed reverse attention to suppress incorrect class predictions in confusing image regions. This suppression effectively redirected the model's attention where it underperforms. Li \etal used a ``reverse-and-distill" strategy to disentangle attribute and object representations, where reverse attention is used to guide the learning of less visible semantic components by masking confident attribute/object features~\cite{li2023distilled}. Although it does not use reverse attention per se, Hyeon-Woo \etal~\cite{hyeon2023scratching} tackle the bias of peaked softmax attention by injecting uniform attention to support denser token interactions. Following these works, RCA indirectly enhances mid-level attention activation by suppressing extremes in the transformer attention matrix, allowing focus redistribution across insufficiently attended yet semantically relevant image tokens.

Our approach is also informed by the hierarchical design by Wang \etal~\cite{wang2024ra}, where stacked attention modules progressively emphasize important image regions. We abstract this concept by applying contrastive modulation not on spatial features, but rather, indirectly to attention weights with a parameter-based flooring of the final layer hidden states (Eq.~\ref{eq:flooring-operation}). Furthermore, akin to the work ``\emph{Self-Attention with Relative Position Representations}"~\cite{shaw2018self}, which adjusts attention based on position, our method promotes a more balanced and structured distribution of attention across tokens, not by relying on positional information (\eg positional encoding) but rather by contrast enhancement to emphasize moderately attended regions and suppress overly dominant or neglected ones. 

Recent works such as RA-Net~\cite{wang2024ra}, RTA-Former~\cite{li2024rta}, and SRaNet~\cite{lee2023shallow} confirm the growing value of reverse attention mechanisms in guiding residual learning, boundary recovery, and transformer attention refinement. Sun \etal ~\cite{sun2019reverse} introduced both reverse and boundary attention units in a residual refinement module to gradually refine road segmentation by focusing on previously missed regions and road edges. Xie \etal~\cite{xie2019image} introduced learnable bidirectional attention maps, including reverse attention, which suppresses known regions so the U-net can focus solely on reconstructing missing parts. Our RCA offers a general-purpose formulation for this class of methods, operating on attention scores through hidden states to improve detection, grounding, and interpretability across vision-language tasks.

A recent study by Venhoff \etal~\cite{venhoff2025visual} introduces a controlled framework for analyzing how a trainable adapter maps visual representations into the feature space of a frozen LLM. With tools based on the sparse autoencoder (SAE), they demonstrate that vision-language alignment predominantly emerges in the middle-to-late transformer layers. This discovery reinforces the rationale behind RCA, which operates directly on the final transformer layer's attention matrix to boost weakly attended but semantically relevant tokens. The SAE-based observations indicate that attentional reweighting strategies like RCA are most effective when applied to the layers where visual features have already begun to resemble the internal language representations, specifically, the mid-to-late layers of the transformer where cross-modal alignment naturally occurs.

\section{Methodology}
\label{sec:methods}
\subsection{Computer Vision Task}

Various VLMs were subjected to the task of identifying and localizing (with bounding boxes) all objects in an image matching a free-form natural language prompt, despite the possible absence of object categories in a fixed label set. We refer to this task as \emph{Open Vocabulary Referring Object Detection} (OV-RefOD), which has recently emerged at the interface between open-vocabulary object detection (OVD) and referring expression comprehension (REC), such as in visual grounding~\cite{wang2024ov} and attribute recognition~\cite{chen2023ovarnet}.  

OV-RefOD is open-ended localization guided by natural language, where:
\begin{itemize}
    \item \textbf{Input:} image and an arbitrary text prompt
    \item \textbf{Output:} bounding box coordinates {\scriptsize\texttt{[x1, y1, x2, y2]}} (parsed) of at least one of all visible instances in the image matching the description
\end{itemize}
The concept behind the task has been gaining traction in discussions around VLM evaluation. Notable developments include the introduction of OV-VG, a benchmark for open-vocabulary visual grounding~\cite{wang2024ov} and phrase localization~\cite{zhou2025led}; GroundVLP~\cite{shen2024groundvlp}, which exploits zero-shot visual grounding from vision-language pretraining and OVD; and grounded spatial reasoning in VLMs~\cite{cheng2025spatialrgpt}, which released the Open Spatial Dataset with five million open-vocabulary boxes and masks to test grounding capabilities. More recent efforts include LED~\cite{zhou2025led}, which augments OV detectors by integrating hidden states from LLMs to enhance grounding on OmniLabel benchmarks, and zPROD~\cite{sinhamahapatra2025zero}, which introduces a zero-shot framework for OVD, segmentation, and grounding in challenging autonomous-driving contexts. Collectively, these studies reflect a broader trend toward systematically evaluating VLM grounding under open-vocabulary and zero-shot regimes. 

\subsection{Reverse Contrast Attention}
\label{sec:rca}
This paper introduces Reverse Contrast Attention (RCA), a novel method inspired by treating the transformer's attention matrix as an image. In this analogy, the attention matrix reveals visual patterns and contrast adjustment can be used to emphasize certain features. Traditional contrast enhancement amplifies extremes, making high values brighter and low values darker, effectively emphasizing dominant patterns. However, reverse contrast enhancement suppresses extremes and brings out mid-range features that might otherwise be overlooked. RCA applies this principle to the final-layer attention maps of VLMs, effectively establishing a ``floor" on the hidden states. This adjustment improves the model's sensitivity to moderately activated but semantically relevant visual tokens, thereby enhancing the average precision of its responses in OV-RefOD tasks.

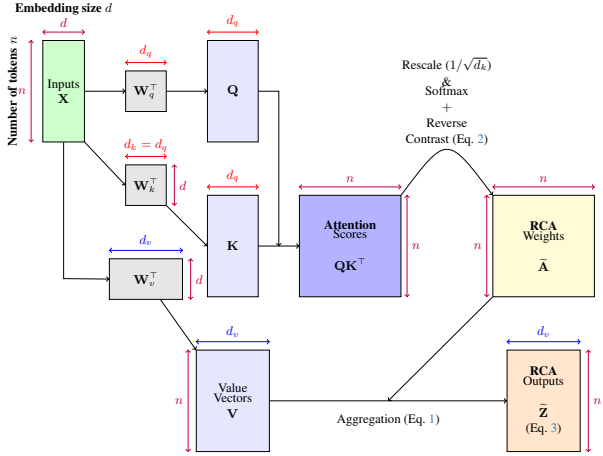
\begin{figure}[htbp]
    \centering
    \scalebox{0.54}{ 
    \begin{tikzpicture}[every node/.style={font=\small}, node distance=1.2cm and 1cm]

        \node[draw, fill=green!20, minimum width=1.0cm, minimum height=2.5cm, align=center] (X) {Inputs \\ $\mathbf{X}$};
        
        \node[right=of X, draw, fill=gray!20, minimum width=1cm, minimum height=1.0cm] (Wq) {$\mathbf{W}_q^\top$};
        \node[below=1.3cm of Wq, draw, fill=gray!20, minimum width=1cm, minimum height=1.0cm] (Wk) {$\mathbf{W}_k^\top$};
        \node[below=1.3cm of Wk, draw, fill=gray!20, minimum width=1.8cm, minimum height=1.0cm] (Wv) {$\mathbf{W}_v^\top$};
        
        \node[right=of Wq, draw, fill=blue!10, minimum width=1.25cm, minimum height=2.5cm] (Q) {$\mathbf{Q}$};
        \node[below=1.3cm of Q, draw, fill=blue!10, minimum width=1.25cm, minimum height=2.5cm] (K) {$\mathbf{K}$};
        \node[below=1.3cm of K, draw, fill=blue!10, minimum width=1.8cm, minimum height=2.5cm, align=center] (V) {\small{Value}\\[-1.0ex]\small{Vectors} \\ $\mathbf{V}$};
        
        \node[right=of K, draw, fill=blue!30, minimum width=2.5cm, minimum height=2.5cm,align=center] (QK) {\small{\textbf{Attention}}\\[-1.0ex]\small{Scores}\\ \\$\mathbf{QK}^\top$};
        \node[right=of QK, draw=none] (softmax) {};
        \node[right=of softmax, draw, fill=yellow!20, minimum width=2.5cm, minimum height=2.5cm, align=center] (A) {\small{\textbf{RCA}}\\[-1.0ex]\small{Weights} \\\\ $\mathbf{\widetilde{A}}$};
        
        \node[below=1.3cm of A, draw, fill=orange!20, minimum width=1.8cm, minimum height=2.5cm, align=center] (Z) {\textbf{RCA}\\[-1.0ex]Outputs \\\\ $\mathbf{\widetilde{Z}}$\\ (Eq.~\ref{eq:flooring-operation})};
        
        \draw[->] (X) -- (Wq);
        \draw[->] (X.south east) -- (Wk.west);
        
        \draw[->] (Wq) -- (Q);
        \draw[->] (Wk.south east) -- (K.west);
        \draw[->] (Wv) -- (V.north west);
        
        \path (Q.east) ++(0.5,0) coordinate (Qout); 
        \path let 
          \p1 = (K.east), 
          \p2 = (QK.west) 
        in coordinate (KtoQKmid) at ($(\p1)!.5!(\p2)$);
        \path let 
          \p1 = (V.east), 
          \p2 = (Z.west) 
        in coordinate (VtoZmid) at ($(\p1)!.5!(\p2)$);

        \path (X.south) ++(0,-3.36) coordinate (Xout);
        
        \draw[->] (X.south) -- (Xout) -- (Wv.west);
        \draw[->] (Q.east) -- (Qout) -- (KtoQKmid);
        \draw[->] (K) -- (QK);
        
        \path let 
          \p1 = (QK.north east),
          \p2 = (A.north west)
        in coordinate (CurveApex) at ($(\p1)!.5!(\p2) + (0,1.5)$);
        
        \draw[->] (QK.north east) .. controls (CurveApex) .. (A.north west);
        
        \draw[->] (A.south west) -- (VtoZmid);
        \draw[->] (V) -- (Z);
        
        \node[above=0.5cm of X] (xlab) {\textbf{Embedding size $d$}};
        \node[left=0.5cm of X] {\rotatebox{90}{\textbf{Number of tokens $n$}}};
        \node[above=-0.4cm of CurveApex, align=center] {Rescale ($1/\sqrt{d_k}$)\\[-1.5ex]\\[-1.0ex]\&\\[-1.0ex]Softmax\\
        $+$\\Reverse\\Contrast (Eq.~\ref{eq:renormalization})};
        \node[below=0.2cm of VtoZmid] {Aggregation (Eq.~\ref{eq:aggregation-hidden})};
        
        \draw[<->, red] 
          ([yshift=0.2cm]Wq.north east) -- ([yshift=0.2cm]Wq.north west) 
          node[midway, above] {$d_q$};
        \draw[<->, red] 
          ([yshift=0.2cm]Q.north east) -- ([yshift=0.2cm]Q.north west) 
          node[midway, above] {$d_q$};  
        \draw[<->, red] 
          ([yshift=0.2cm]K.north east) -- ([yshift=0.2cm]K.north west) 
          node[midway, above] {$d_q$};  
            
        \draw[<->, red] ([yshift=0.2cm]Wk.north east) -- ([yshift=0.2cm]Wk.north west) node[midway, above] {$d_k = d_q$};
        \draw[<->, blue] ([yshift=0.2cm]Wv.north east) -- ([yshift=0.2cm]Wv.north west) node[midway, above] {$d_v$};
        \draw[<->, blue] ([yshift=0.2cm]V.north east) -- ([yshift=0.2cm]V.north west) node[midway, above] {$d_v$};
        \draw[<->, blue] ([yshift=0.2cm]Z.north east) -- ([yshift=0.2cm]Z.north west) node[midway, above] {$d_v$};
        
        \draw[<->, purple] ([yshift=0.2cm]X.north east) -- ([yshift=0.2cm]X.north west) node[midway, above] {$d$};
        \draw[<->, purple] ([xshift=1.1cm]Wv.north) -- ([xshift=1.1cm]Wv.south) node[midway, right] {$d$};
        \draw[<->, purple] ([xshift=0.7cm]Wk.north) -- ([xshift=0.7cm]Wk.south) node[midway, right] {$d$};

        \draw[<->, purple] ([xshift=-0.8cm]X.north) -- ([xshift=-0.8cm]X.south) node[midway, left] {$n$};
        \draw[<->, purple] ([xshift=-1.1cm]V.north) -- ([xshift=-1.1cm]V.south) node[midway, left] {$n$};
        \draw[<->, purple] ([yshift=0.2cm]QK.north east) -- ([yshift=0.2cm]QK.north west) node[midway, above] {$n$};
        \draw[<->, purple] ([xshift=1.4cm]QK.north) -- ([xshift=1.4cm]QK.south) node[midway, right] {$n$};
        \draw[<->, purple] ([xshift=1.1cm]Z.north) -- ([xshift=1.1cm]Z.south) node[midway, right] {$n$};
        \draw[<->, purple] ([xshift=-1.4cm]A.north) -- ([xshift=-1.4cm]A.south) node[midway, left] {$n$};
        \draw[<->, purple] ([yshift=0.2cm]A.north east) -- ([yshift=0.2cm]A.north west) node[midway, above] {$n$};
                
    \end{tikzpicture}
    }
    \caption{Illustration of the RCA mechanism.}
    \label{fig:rca_attention}
\end{figure}

In a standard transformer layer, the hidden state $z_i$ is a vector that represents the token $i$ as a superposition of the value vectors $v_j\in\mathbf{V}$ of the text prompt and the image regions according to the attention distribution $\mathbf{A}=\left\{\alpha_{ij}\right\}$ (Figure~\ref{fig:rca_attention}): 
\begin{equation}
\label{eq:aggregation-hidden}    
z_i = \sum_{j=1}^n \alpha_{ij}v_j \in \mathbf{Z}.
\end{equation}

RCA restructures $\mathbf{A}$ by amplifying $\alpha_{ij}$ that are near some central value $m$, while inhibiting $\alpha_{ij}$ that are far above or below $m$. This restructuring can be accomplished by a nonmonotonic reweighting that can particularly take either form:
\begin{itemize}[leftmargin=*]
    \item \textbf{Inverse distance from $m$:}
    \begin{align*}
        \alpha'_{ij} = \frac{1}{1 + \gamma \left| \alpha_{ij} - m \right|}
    \end{align*}
    \item \textbf{Gaussian peaking around $m$:}
    \begin{align*}
        \alpha'_{ij} = \exp\left( -\gamma \left[ \alpha_{ij} - m \right]^2 \right)
    \end{align*}
\end{itemize}
The free, but possibly optimizable, parameter $\gamma$ regulates \emph{reverse contrast}, or how sharply large deviations from $m$ are penalized. Without loss of generality, we take $\gamma=1$ whereas $m$, which is another potentially optimizable free parameter, is manually selected.

The reverse contrast weights $\alpha'$ then go through a renormalization:
\begin{equation}
    \label{eq:renormalization}
    \widetilde{\alpha_{ij}} = \frac{\max\left(\alpha'_{ij},0\right)}{\sum_{k=1}^n\max\left(\alpha'_{ij},0\right)},
\end{equation}
which are $\in\mathbf{\widetilde{A}}$ in Figure~\ref{fig:rca_attention}. We claim that RCA implies the element-wise flooring operation applied to the final-layer hidden states such that:
\begin{equation}
    \label{eq:flooring-operation}
    \widetilde{\mathbf{Z}} = \left\{\max\left(z_i,\vartheta\right)| z_i\in \mathbf{Z}, \vartheta\in\mathbb{R}\right\},
\end{equation}
in which $\vartheta$ is a free parameter linked to $m$ and $\gamma$. Adjusting $\vartheta$ implies tuning of $m$ or $\gamma$.

\subsection{Inference}
During inference, a model receives an input pair, $\mathbf{X}=(I,Q)$, where $I$ is an image, and $Q$ is the free-form natural language query:
\begin{quote}
    \scriptsize
    \texttt{'Give the normalized bounding box coordinates in the format [x1, y1, x2, y2] of all instances of \{cls\} in the image.'}
\end{quote}
in which \texttt{\{cls\}} refers to the category or the descriptive phrase of the object, and \texttt{x1}, \texttt{x2}, \texttt{y1}, and \texttt{y2} are ideally floating point values $\in[0,1]$. This prompt was applied in all VLM to generate one or more parsed bounding boxes $\left\{B_k\right\}$ that supposedly align with the object(s) referenced in $Q$. 
Some models returned $B_k$ in pixel coordinates rather than in normalized format. To ensure consistency, all outputs were standardized to a common format using regular expression parsing.

\subsection{VLM Selection}
We considered open-source VLMs from the OpenCompass Multi-Modal Academic Leaderboard (OC-MMAL) at \href{https://rank.opencompass.org.cn/leaderboard-multimodal}{https://rank.opencompass.org.cn/leaderboard-multimodal} due to their publicly benchmarked performance on image-text reasoning tasks and other multi-modal capabilities. The VLM must satisfy the following criteria:
\begin{itemize}
    \item LLM size $< 35$B parameters, for efficiency reasons;
    \item Provides bounding box coordinates in the form $B_k$;
    \item Checkpoints available in HuggingFace
\end{itemize}
The list samples top- and middle-ranking open-source VLM in OC-MMAL:
\begin{table}[htbp]
    \centering
    \scriptsize
    \begin{tabular}{p{2.9cm}|p{0.54cm}|p{1.5cm}|p{1.5cm}}\hline
        \multicolumn{1}{c|}{Model} &  Params & \multicolumn{1}{c|}{LLM} & \multicolumn{1}{c}{Vision} \\\hline
        \texttt{Ovis2-34B} & 34.9B & \texttt{Qwen2.5-32B}&\texttt{AIMv2-1B}\\
        \texttt{SAIL-VL-1.6-8B} & 8.33B &\texttt{Qwen2.5-7B}&\texttt{AIMv2 Huge}\\
        \texttt{WeThink-Qwen2.5VL-7B} & 8.29B &\texttt{Qwen2.5-7B} &\texttt{QwenViT}\\
        \texttt{Qwen2.5-VL-7B} & 8.29B &\texttt{Qwen2.5-7B} &\texttt{QwenViT}\\
        \texttt{MiniCPM-o-2.6} & 8.67B &\texttt{Qwen2.5-7B}&\texttt{SigLIP-400M}\\
        \texttt{valley2\_dpo} & 8.88B &\texttt{Qwen2.5-7B} &\texttt{SigLIP-400M}\\
        \texttt{Kimi-VL-A3B} & 16.4B & \texttt{Moonlight}\newline\texttt{-16B-A3B} & \texttt{MoonViT}\\
        \texttt{Ristretto-3B} & 3.84B & \texttt{Qwen2.5-3B} &\texttt{SigLIP-400M}\\
        \texttt{POINTS1.5-Qwen2.5-7B} & 8.3B &\texttt{Qwen-2.5-7B} &\texttt{NaViT}\\
        \texttt{Valley-Eagle} & 8.9B &\texttt{Qwen2.5-7B} &\texttt{SigLIP-400M}\\
        \texttt{Gemma3-27B}& 27.4B &\texttt{Gemma3-27B}&\texttt{SigLIP-400M}\\
        \texttt{VARCO-VISION-14B}& 15.2B &\texttt{Qwen2.5-14B}&\texttt{SigLIP-400M}\\
        \texttt{DeepSeek-VL2}& 27.5B &\texttt{DeepSeekMoE}\newline\texttt{-27B} &\texttt{SigLIP-400M}\\
        \texttt{PaliGemma2-3B-mix-448}&3B&\texttt{Gemma2-2B}&\texttt{SigLIP-400M}\\
        \texttt{Moondream2} & 1.9B&\texttt{Phi-1.5}&\texttt{SigLIP-400M}\\
        \hline
    \end{tabular}
    \caption{Selected VLMs from OC-MMAL in ascending rank.}
    \label{tab:VLM_selection}
\end{table}
Although we attempted as broad a coverage as possible, some models like \texttt{Ola-7b} and \texttt{Intern-VL} do not respond to the prompt $Q$ with parsable bounding box information or were designed to interpret the prompt as a REC rather than OV-RefOD. Other models are earlier versions or have a version with a higher rank in OC-MMAL; hence, we opt for the updated or higher ranked version.

\subsection{Evaluation}
In this paper, we propose \textbf{FitAP} (supplementary section~\ref{section:fitap}), a modified evaluation metric derived from Average Precision (AP), commonly used in object detection tasks~\cite{everingham2010pascal}. Unlike standard AP, which ranks detections by confidence scores typically produced by region proposal networks~\cite{ren2015faster}, FitAP is designed to evaluate OV-RefOD in VLMs that lack explicit confidence output. FitAP ranks predicted bounding boxes according to the product of their normalized area $A_{\text{box}}$ and their intersection-over-union (IoU) with ground-truth annotations. This alternative ranking strategy preserves the precision-recall structure of AP while enabling evaluation in settings where traditional confidence-based sorting is not available or unreliable. Thus, FitAP provides a practical and interpretable measure for assessing detection quality in VLMs operating under weakly supervised or prompt-based regimes~\cite{kamath2021mdetr, li2022grounded}.

We evaluated the models in Table~\ref{tab:VLM_selection} on the novel split of \texttt{COCO val 2017}~\cite{zareian2021open,gu2022openvocabulary} consisting of $2064$ $(I,Q)$ pairs. The mean FitAP is the average value at different IoU thresholds $\in[0.5:0.05:0.95]$.

\section{Results and Discussion}
\begin{table}[htbp]
    \centering
    \scriptsize
    \begin{tabular}{p{2.9cm}|p{1.0cm}|p{1.0cm}|r}
        \hline
        \multicolumn{1}{c|}{} & \multicolumn{2}{c|}{\textbf{FitAP} ($\mathbf{\uparrow}$)} & \\
        \cline{2-3}
        \multicolumn{1}{c|}{Model} & pre-RCA & post-RCA & \% Change \\
        \hline
        \texttt{Ovis2-34B} & $3.23869$ & $3.52222$ & $\mathbf{+8.75}$ \\
        \texttt{SAIL-VL-1.6-8B} & $4.84873$ & $5.67149$ & $\mathbf{+17.0}$ \\
        \texttt{WeThink-Qwen2.5VL-7B} & $39.9640$ & $37.7606$ & $-5.51$ \\
        \texttt{Qwen2.5-VL-7B} & $37.0005$ & $46.8535$ & $\mathbf{+26.6}$ \\
        \texttt{MiniCPM-o-2.6} & $0.03064$ & $0.07334$ & $\mathbf{+139}$ \\
        \texttt{valley2\_dpo} & $11.5145$ & $11.6927$ & $\mathbf{+1.55}$ \\
        \texttt{Kimi-VL-A3B} & $30.7194$ & $32.2176$ & $\mathbf{+4.88}$ \\
        \texttt{Ristretto-3B} & $9.12887$ & $7.94552$ & $-13.0$ \\
        \texttt{POINTS1.5-Qwen2.5-7B} & $9.75203$ & $9.45686$ & $-3.03$ \\
        \texttt{Valley-Eagle} & $11.7736$ & $11.2598$ & $-4.36$ \\
        \texttt{Gemma3-27B} & $2.74179$ & $3.01913$ & $\mathbf{+10.1}$ \\
        \texttt{VARCO-VISION-14B} & $27.3592$ & $28.7003$ & $\mathbf{+4.90}$ \\
        \texttt{DeepSeek-VL2} & $3.38530$ & $3.99586$ & $\mathbf{+18.0}$ \\
        \texttt{PaliGemma2-3B-mix-448} & $38.7982$ & $41.1179$ & $\mathbf{+5.98}$ \\
        \texttt{Moondream2} & $47.0039$ & $47.0819$ & $\mathbf{+0.17}$ \\
        \hline
    \end{tabular}
    \caption{FitAP of VLM before and after applying RCA.}
    \label{tab:RCA_effect}
\end{table}
RCA improved six of the top seven and, in general, 11 of 15 VLM in Table~\ref{tab:VLM_selection}, despite the absence of systematic optimization of the $\vartheta$ parameter in Eq.~\ref{eq:flooring-operation}. The authors of \texttt{WeThink-Qwen2.5VL-7B}~\cite{yang2025wethink} observed that additional training stages, especially supervised fine-tuning and chain-of-thought, degraded grounding precision and latent object detection capability of \texttt{Qwen2.5-VL-7B}, which could partially explain the negative effect of RCA on this model. For \texttt{MiniCPM-o-2.6}, the low FitAP is probably due to internal randomization imposed on response generation through sampling decoding~\cite{yu2025rlaif}. 

In the bottom half of the list (Table~\ref{tab:RCA_effect}) is \texttt{PaliGemma2-3B-mix-448}, which, like \texttt{Gemma3-27B} garnered a positive RCA effect. The more significant OV-RefOD improvement of \texttt{PaliGemma2-3B-mix-448} than \texttt{Gemma3-27B} can be explained by its integration of image and text at the model level, enabling deep cross-modal reasoning; hence, naturally suitable for object detection and visual grounding. \texttt{Gemma3} entirely lacks this modality fusion.

\begin{figure*}[htbp]
  \centering
  \begin{subfigure}{\linewidth}
    \includegraphics[trim=20 220 141 141,clip,width=\linewidth]{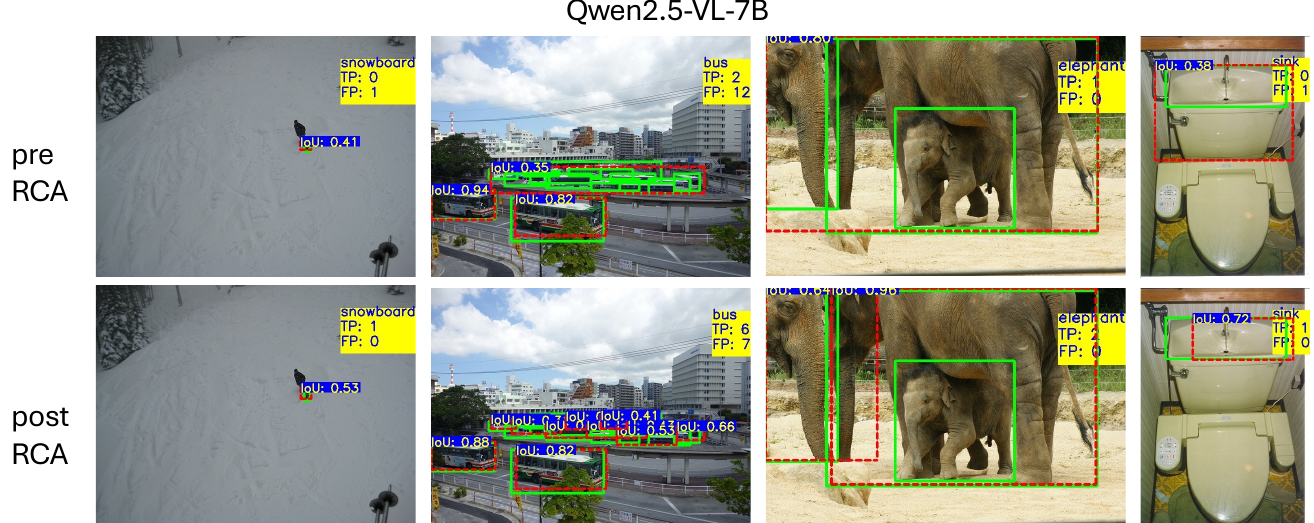}
  \end{subfigure}
    \caption{Selected examples suggestive of RCA's positive impact on \texttt{Qwen2.5-VL-7B}: solid, green boxes (ground truth); dashed, red boxes (parsed detections). These cases illustrate improved object localization and precision following the application of RCA.}
    \label{fig:qwen-results}
\end{figure*}

\begin{figure*}[htbp]
  \centering
  \begin{subfigure}{\linewidth}
    \includegraphics[trim=20 160 141 141,clip,width=\linewidth]{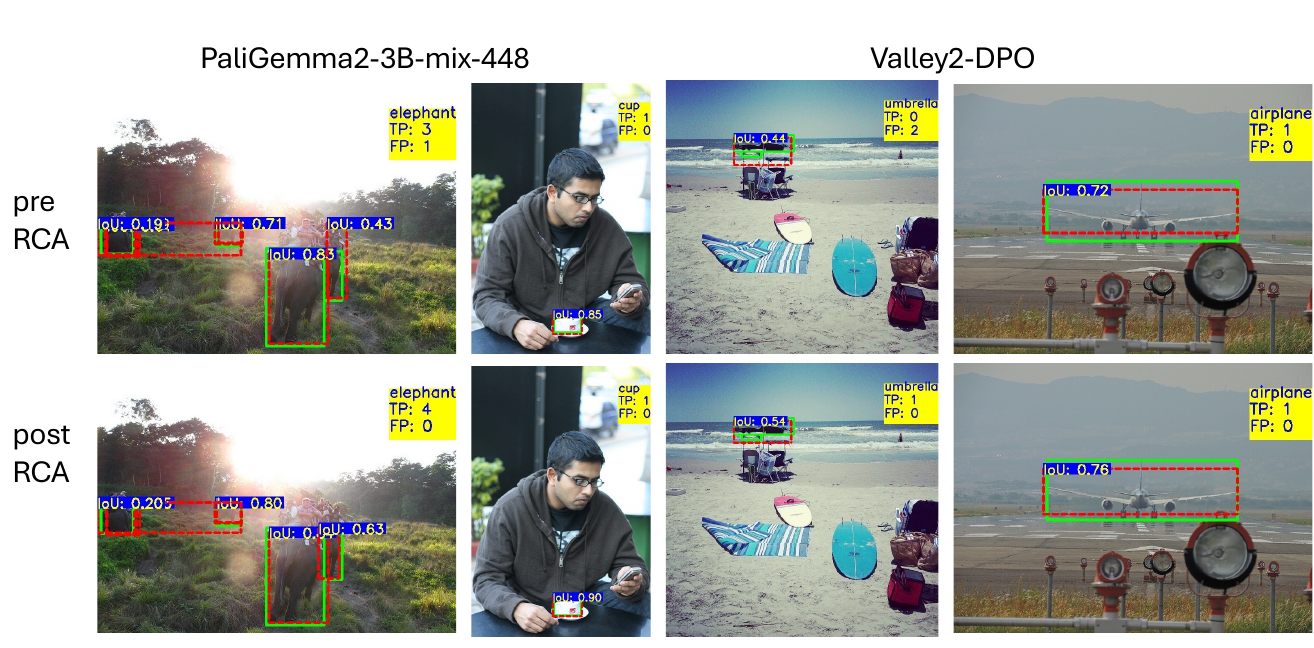}
  \end{subfigure}
    \caption{Selected examples from \texttt{PaliGemma2-3B-mix-448} and \texttt{valley2\_dpo}, qualitatively illustrating the observed improvements in detection after RCA is applied: solid green boxes (ground truth); dashed red boxes (detections)}
    \label{fig:paligemma-valley2-results}
\end{figure*}

\begin{figure*}[htbp]
    \centering
    \begin{subfigure}{\linewidth}        
        \includegraphics[trim=60 130 31 134,clip,width=\linewidth]{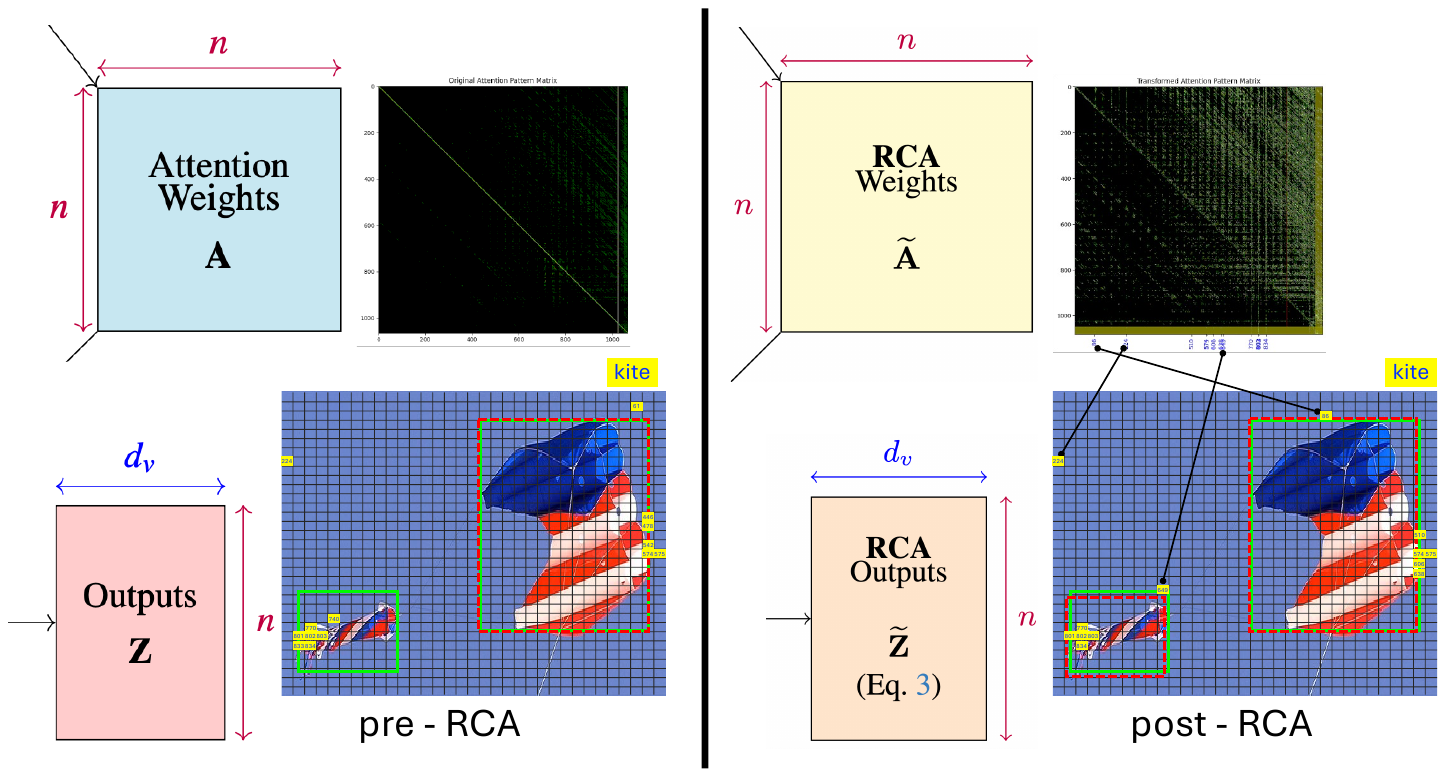}
    \end{subfigure}
        \caption{Visualizing the link between RCA and OV-RefOD in \texttt{PaliGemma2-3B-mix-448} for a sample of \texttt{kite} from \texttt{COCO val 2017} base vocabulary subset: (\emph{left}) before applying RCA; (\emph{right}) after applying RCA. Attention matrix images are beside their corresponding diagram; and patched image directly below the corresponding attention matrix image. The outputs generate bounding box information that can be drawn on the image. Solid green boxes are ground truths; dashed red boxes are the parsed VLM detections of the class \texttt{kite}. The highlighted patches (yellow with blue font) correspond to sufficiently attended image tokens. The indexes of these tokens are shown in the post-RCA attention matrix. Some patches that emerged post-RCA are linked to their corresponding positions in the image for emphasis.}
        \label{fig:visual-rca}
\end{figure*}

Figures~\ref{fig:qwen-results} and~\ref{fig:paligemma-valley2-results}, in which TP and FP represent true positive and false positive detection, respectively, illustrate the improved recall and precision that corroborate positive FitAP changes due to RCA (Table~\ref{tab:RCA_effect}). Green boxes (solid) belong to ground-truth annotations from \texttt{COCO val 2017}, while red boxes (dashed) correspond to parsed VLM detections. In Fig.~\ref{fig:qwen-results}, RCA enhanced the detection of multiple instances of objects $\texttt{bus}$ and \texttt{elephant}, while increasing its precision in detecting small objects such as \texttt{snowboard} and \texttt{sink} (higher IoU). In Fig.~\ref{fig:paligemma-valley2-results}, RCA sharpened box precision leading to higher IoU in multiple instances of \texttt{elephant}, and single instances of \texttt{cup}, \texttt{umbrella}, and \texttt{airplane} in their respective images. These improvements are explainable with RCA enforcing a shift in attention focus toward subdued image tokens (Figure~\ref{fig:visual-rca}).



In the original attention matrix (Fig.~\ref{fig:visual-rca}, left), the highlighted patch indices exhibit strong attention scores that extend downward to the final rows. However, several tokens corresponding to distinct image regions remain visually indistinct, suggesting insufficient attention. After applying RCA, these previously subdued tokens are amplified in the transformed attention matrix (right), making their associated image patches more prominent. This redistribution of attention likely allowed the model to detect an additional object instance, namely the \texttt{kite}, which was not distinguished in the pre-RCA outputs.

The RCA propagates its influence through a structured transformation of $\mathbf{A}$ centered on a chosen mid-value $m$, which is the \emph{mean of the column-wise (i.e., along the $\mathbf{K}$-dimension of $\mathbf{A}$) maximum attention across multiple heads}. This restructuring inhibits extremes, that is, tokens or image regions receiving disproportionately high or low attention, while it amplifies contributions closer to the midpoint. Thus, $\widetilde{\mathbf{A}}$ is a more equalized attention profile, implying more stable and bounded hidden states from the final transformer layer. As extreme attentions are suppressed, the resultant vectors from applying the superposition (Eq.~\ref{eq:aggregation-hidden}) on $\widetilde{\mathbf{A}}$ rather than on $\mathbf{A}$ are less likely to fall below a threshold $\vartheta$ (supplementary Fig.~\ref{fig:flooring-example}) as outlined in the following argument. Suppose that
\begin{equation}
\label{eq:superposition-rca-states}
\widetilde{z_i}=\sum_{j=1}^n\widetilde{\alpha_{ij}}v_j
\end{equation}
represents the transformed hidden states from the final transformer layer, where $\widetilde{\alpha_{ij}}\in\widetilde{A}$ is the renormalized attention weight (Eq.~\ref{eq:renormalization}) for the query token $i$ over ``key" token $j$, and $v_j\in\mathbb{R}^{d_v\times 1}$ is the value vector for the token $j$. Consider in further detail a partition of tokens into two sets of disjoint token index.
\begin{itemize}[leftmargin=*]
    \item \textbf{Subthreshold contributors:} are tokens that dip below $\vartheta$ in dimension $d$.
    \begin{align*}
        \mathcal{J}_{\downarrow} = \left\{j|v_j(d)<\vartheta\right\}
    \end{align*}
    \item \textbf{Suprathreshold contributors:} are tokens of which value vector is at least $\vartheta$ in dimension $d$.
    \begin{align*}
        \mathcal{J}_{\uparrow} = \left\{j|v_j(d)\geq\vartheta\right\}
    \end{align*}
\end{itemize}
where the inequalities hold element-wise in all dimensions. Applying this partition to each component $d$, which is the $d^{\text{th}}$ scalar element representing one channel of the model's internal feature space, in Eq.~\ref{eq:superposition-rca-states}, then 
\begin{equation}
    \label{eq:partitioned-superposition-rca-states}
    \widetilde{z_i}(d) = \sum_{j\in\mathcal{J}_{\downarrow}}\widetilde{\alpha_{ij}}v_j(d) + \sum_{j\in\mathcal{J}_{\uparrow}}\widetilde{\alpha_{ij}}v_j(d)
\end{equation}

Performing the partition (Section~\ref{sec:guarantee-rca-flooring}) ultimately leads to the following inequality:
\begin{equation}
    \label{ineq:formal-flooring-inequality}
    \widetilde{z_i}(d) \geq  \vartheta + \underset{\textup{negative}}{\underbrace{\left(v^- - \vartheta\right)}}\underset{\textup{small if penalized}}{\underbrace{\left(\sum_{j\in\mathcal{J}_{\downarrow}} \widetilde{\alpha_{ij}}\right)}} .
\end{equation}
When the penalty term is sufficiently small, $\widetilde{z_i}(d)$ ends up ``close enough" to $\vartheta$, resulting in the RCA outputs defined in Eq.~\ref{eq:flooring-operation}. The term $\sum_{j\in\mathcal{J}_{\downarrow}}\widetilde{\alpha_{ij}}$, the renormalized mass assigned to the subthreshold contributors, must be small. The guarantee that RCA increases performance relies on this term being close to zero, which could explain why it improved OV-RefOD performance in some models but not in others (Table~\ref{tab:RCA_effect}).

\subsection{Soft Guarantee to RCA Flooring}
\label{sec:guarantee-rca-flooring}
Assume that among the subthreshold contributors to the hidden state of the final transformer layer, the minimal (or near-minimal) component is $v^- < \vartheta$. Then, for $j\in\mathcal{J}_{\downarrow}$, $v_j(d)\geq v^-$, while for $j\in\mathcal{J}_{\uparrow}$, $v_j(d)\geq\vartheta$. Hence, Eq.~\ref{eq:partitioned-superposition-rca-states} simplifies into an inequality:
\begin{equation*}
    \widetilde{z_i}(d)\geq\sum_{j\in\mathcal{J}_{\uparrow}}\widetilde{\alpha_{ij}}\vartheta + \sum_{j\in\mathcal{J}_{\downarrow}}\widetilde{\alpha_{ij}}v^-.
\end{equation*}
Rewrite by factoring out constants:
\begin{equation*}
    \widetilde{z_i}(d)\geq\vartheta\left(\sum_{j\in\mathcal{J}_{\uparrow}}\widetilde{\alpha_{ij}}\right) + v^-\left(\sum_{j\in\mathcal{J}_{\downarrow}}\widetilde{\alpha_{ij}}\right).
\end{equation*}
But by virtue of normalization,
\[
\sum_{j\in\mathcal{J}_{\uparrow}}\widetilde{\alpha_{ij}}+\sum_{j\in\mathcal{J}_{\downarrow}}\widetilde{\alpha_{ij}}=1,
\] 
such that the preceding inequality becomes
\begin{equation*}
    \widetilde{z_i}(d)\geq\vartheta\left(1-\sum_{j\in\mathcal{J}_{\downarrow}}\widetilde{\alpha_{ij}}\right)+v^-\left(\sum_{j\in\mathcal{J}_{\downarrow}}\widetilde{\alpha_{ij}}\right).
\end{equation*}
Thus,
\begin{equation*}
    \widetilde{z_i}(d)\geq \vartheta +\left(v^--\vartheta\right)\left(\sum_{j\in\mathcal{J}_{\downarrow}}\widetilde{\alpha_{ij}}\right),
\end{equation*}
which is the inequality~(\ref{ineq:formal-flooring-inequality}).

If $\sum_{j\in\mathcal{J}_{\downarrow}}\widetilde{\alpha_{ij}}$ is very small, say $\delta\ll 1$, then
\begin{equation*}
    \widetilde{z_i}(d)\geq \vartheta+\left(v^--\vartheta\right)\delta \, \approx \, \vartheta.
\end{equation*}
Of course, $\widetilde{z_i}\geq\vartheta$ is not a strict guarantee, but is based on the original distribution $\mathbf{A}$ of attention weights, the central value $m$, and the subthreshold range, $\left(v^--\vartheta\right)$. The parameter $\vartheta$ could possibly be derived from $\gamma$ and $m$ and could be optimized based on the magnitude of the improvement in the VLM performance in OV-RefOD due to RCA.

\subsection{Empirical Test of RCA Effect}
To empirically verify the assumption underpinning condition~(\ref{ineq:formal-flooring-inequality}), we investigated the correlation between the mean attention weights (a suitable central value) and the number of subthreshold contributions. Specifically, we defined the index set $S=\left\{i: \widetilde{z_i}<\vartheta\right\}$, the cardinality $|S|$ of which represents the number of subthreshold contributions. The mean attention weight is denoted as $m$.

\begin{figure*}[htbp]
  \centering
  \begin{subfigure}{0.33\linewidth}
    \includegraphics[width=\linewidth]{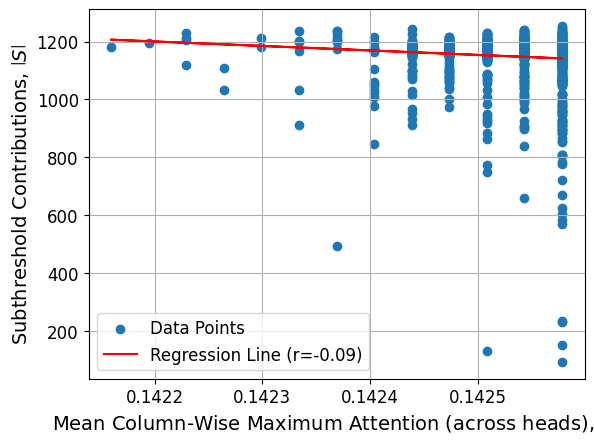}
    \caption{\texttt{Qwen2.5-VL-7B} ($\vartheta=-1.5$)}
  \end{subfigure}
   \begin{subfigure}{0.33\linewidth}
     \includegraphics[width=\linewidth]{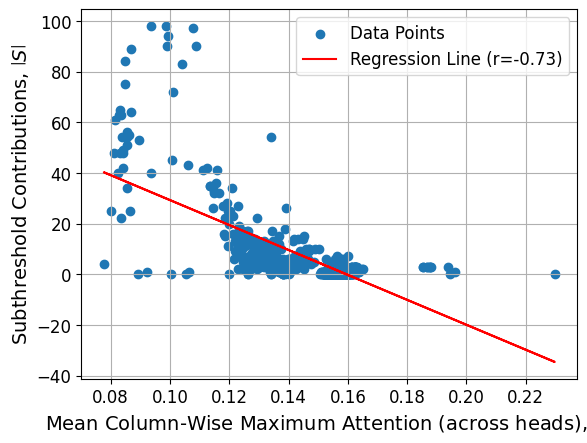}
     \caption{\texttt{DeepSeek-VL2} ($\vartheta=-13$)}
   \end{subfigure}
   \begin{subfigure}{0.33\linewidth}
     \includegraphics[width=\linewidth]{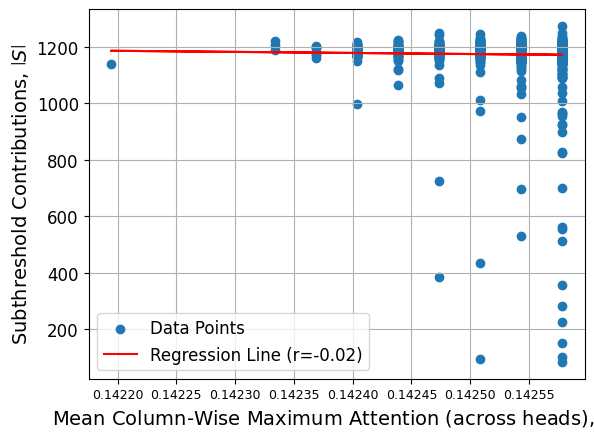}
     \caption{\texttt{WeThink} ($\vartheta=-1.5$)}
   \end{subfigure}
   
    \caption{Correlations between the number $|S|$ of subthreshold contributions and mean cross-head max $m$ of attention weights evaluated on $2064$ $(I,Q)$ pairs with Pearson correlations and $p$-value: (a) $r=-0.09$, $p=0.00004$, (b) $r=-0.73$, $p<0.00001$, (c) $r=-0.02$, $p=0.32$. At the 0.05 confidence level, the correlation coefficient in (c) is not statistically significant.}
    \label{fig:correlation-results}
\end{figure*}

Figure~\ref{fig:correlation-results} plots the $|S|$ against $m$ for \texttt{Qwen2.5-VL-7B}, \texttt{DeepSeek-VL2}, and \texttt{WeThink}. Each data point represents the VLM response to $Q$ on \texttt{COCO val 2017} (all categories included). In the first two cases, a moderately negative but statistically significant Pearson correlation was observed ($r=-0.09$ and $r=-0.73$, respectively). This consistent inverse relationship expresses that when the central value $m$ is lower (that is, attention is more diffusely allocated across tokens), the subthreshold contributions are higher in number. However, the correlation coefficient $r=-0.02$ in \texttt{WeThink} is not significant, corroborating its non-improvement with RCA (Table~\ref{tab:RCA_effect}). 

These observations empirically support the underlying assumption of Eq.~\ref{eq:flooring-operation} that the subthreshold factor found in condition~(\ref{ineq:formal-flooring-inequality}) diminishes with increasing attention sharpness (supplementary section~\ref{section:RCA-correlations}). Hence, by elevating the relevant token attention in these models, the RCA mechanism suppresses the contribution of low-activation (subthreshold) tokens in the final transformer layer. The negative correlation coefficient
supports the hypothesis that RCA enhances model performance by reducing the influence of tokens receiving low attention, effectively reducing noise in the final hidden representations. This selective emphasis endows the model with more precise object localization and response generation.

These theoretical conditions align with the findings of Venhoff \etal~\cite{venhoff2025visual}, who showed that meaningful alignment of vision and language modalities only emerges in the middle to late layers of the transformer. Their use of SAEs revealed that visual tokens become semantically integrated within the LLM's internal representation only in later layers, which is precisely where RCA operates. This observation supports RCA's core premise: modifying attention in late layers is most effective, as the visual features have already been mapped into language-relevant embeddings. In particular, even early-fusion models such as \texttt{DeepSeek-VL2} can benefit from RCA, likely because their high-capacity architectures allow latent modality disentanglement to emerge in deeper layers. In such cases, RCA amplifies meaningful but under-attended visual cues while suppressing irrelevant or overly dominant ones, sharpening the model's focus during object localization.


\subsection{Impact of Vision-Language Fusion Timing}

\begin{figure}[htbp]
    \centering
    \includegraphics[trim=140 150 161 110,clip,width=0.88\linewidth]{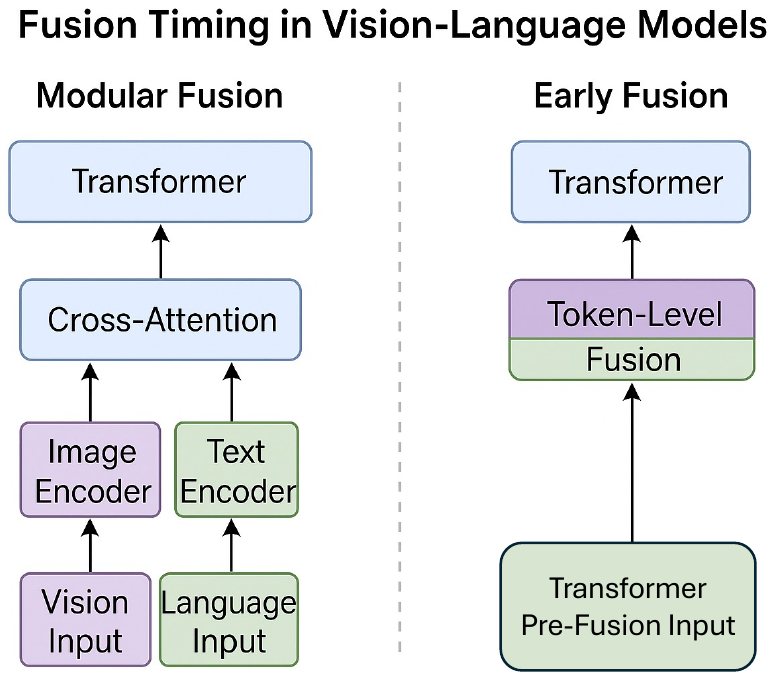}
    \caption{Modular vs. early fusion strategies in VLM}
    \label{fig:modality-fusion}
\end{figure}

Our empirical findings highlight a clear pattern: the \emph{timing and structure of modality fusion} in transformer-based VLMs (Figure~\ref{fig:modality-fusion}) critically determine whether RCA enhances or degrades OV-RefOD performance. The theoretical foundation of RCA (condition~\ref{ineq:formal-flooring-inequality}) requires the suppression of subthreshold visual tokens, which is only possible if visual and linguistic information remain sufficiently separable in the attention mechanism. 

While most models that improved with RCA, such as \texttt{Qwen2.5-VL-7B}, \texttt{SAIL-VL-1.6-8B}, \texttt{MiniCPM-o-2.6}, \texttt{Gemma3-27B}, and \texttt{PaliGemma2-3B}, share a modular late-stage fusion architecture (Fig.~\ref{fig:modality-fusion}), the case of \texttt{DeepSeek-VL2} presents a notable counterexample. Despite employing a tight early fusion strategy using Mixture-of-Experts across modalities, \texttt{DeepSeek-VL2} still benefited from RCA ($+18.0\%$ FitAP). This outcome suggests that tight early fusion is not inherently incompatible with RCA, as long as the model has compensating properties, such as large representational capacity, redundancy-aware design, or latent gating structures that allow attention distributions to evolve separately across layers. This observation is contrary to previous findings that advocate delayed or bottlenecked cross-modal fusion to maintain modality separability and improve alignment and interpretability~\cite{hori2017attention,liang2021attention}. In particular, Hori \etal~\cite{hori2017attention} demonstrate that selective application of attention between modalities during decoding improves output quality, while Liang \etal~\cite{liang2021attention} propose delaying fusion until each modality is internally encoded to minimize distributional mismatch. \texttt{DeepSeek-VL2}'s RCA compatibility suggests that learned internal disentanglement, enabled by high-capacity attention pathways, may offer an alternative path to attention reweighting success, even in early fusion architectures.


In contrast, models that degraded post-RCA, such as \texttt{Valley-Eagle}, \texttt{WeThink}, and \texttt{POINTS1.5}, share characteristics that limit the effectiveness of RCA. These include early token-level fusion without structural constraints or training objectives, such as Chain-of-Thought supervision or instruction tuning, that emphasize abstract reasoning over spatial grounding. In the case of \texttt{Valley-Eagle}, for example, the Eagle module directly integrates image tokens into the transformer embedding space early in the architecture~\cite{wu2025valley2}. \texttt{WeThink-Qwen2.5-VL}, while built on a strong modular foundation, was subject to reinforcement learning and chain of thought supervision that deemphasized precise visual grounding~\cite{yang2025wethink}. \texttt{POINTS1.5}, similarly, adopts a modality fusion scheme designed for efficiency and general reasoning rather than for attention interpretability and separability. In these cases, the distribution of attention on the visual tokens remains too diffuse or semantically entangled, violating the assumption in condition~(\ref{ineq:formal-flooring-inequality}) that subthreshold contributors can be isolated and suppressed (Fig.~\ref{fig:flooring-example}). These findings reinforce the need to consider the fusion strategy when designing VLMs for explainability or plug-in interpretability methods such as RCA. Models with sufficient capacity and flexible attention dynamics can still benefit from RCA even with an early stage fusion strategy, as demonstrated by \texttt{DeepSeek-VL2}.

While we focused on architectural factors to explain RCA’s effectiveness, other elements—like dataset composition, prompt structure, and implementation choices—may also influence outcomes. These factors could interact with architecture in subtle ways, warranting future research to explore their interplay and uncover broader principles behind attention-based inference-time interventions.


\section{Conclusion}
This work proposed Reverse Contrast Attention (RCA), a novel method that reformulates attention maps at inference time to enhance object localization in vision-language transformers without altering model parameters. Through both theoretical analysis and empirical evaluation on the OV-RefOD task, we demonstrate that RCA selectively boosts the influence of semantically relevant but neglected tokens, thereby improving interpretability and performance. The introduction of FitAP allows meaningful benchmarking in the absence of explicit confidence scores, and our findings highlight the importance of architectural factors such as late-stage modality fusion to make RCA effective. Beyond performance gains, RCA offers a diagnostic lens into the internal workings of VLMs, showing that attention plasticity, when deliberately guided, can serve as a tool not only for eXCV but for functional enhancement. These insights lay the groundwork for future research into adaptive attention reweighting and post hoc interpretability methods across multimodal models.
{
    \small
    \bibliographystyle{ieeenat_fullname}
    \bibliography{main}

\begin{thebibliography}{32}
\providecommand{\natexlab}[1]{#1}
\providecommand{\url}[1]{\texttt{#1}}
\expandafter\ifx\csname urlstyle\endcsname\relax
  \providecommand{\doi}[1]{doi: #1}\else
  \providecommand{\doi}{doi: \begingroup \urlstyle{rm}\Url}\fi

\bibitem[Chefer et~al.(2021{\natexlab{a}})Chefer, Gur, and Wolf]{chefer2021generic}
Hila Chefer, Shir Gur, and Lior Wolf.
\newblock Generic attention-model explainability for interpreting bi-modal and encoder-decoder transformers.
\newblock In \emph{ICCV}, pages 397--406, 2021{\natexlab{a}}.

\bibitem[Chefer et~al.(2021{\natexlab{b}})Chefer, Gur, and Wolf]{chefer2021transformer}
Hila Chefer, Shir Gur, and Lior Wolf.
\newblock Transformer interpretability beyond attention visualization.
\newblock In \emph{CVPR}, pages 782--791, 2021{\natexlab{b}}.

\bibitem[Chen et~al.(2023)Chen, Jiang, Hu, Tang, Gao, Chen, and Xie]{chen2023ovarnet}
Keyan Chen, Xiaolong Jiang, Yao Hu, Xu Tang, Yan Gao, Jianqi Chen, and Weidi Xie.
\newblock Ovar{N}et: Towards open-vocabulary object attribute recognition.
\newblock In \emph{CVPR}, pages 23518--23527, 2023.

\bibitem[Chen et~al.(2018)Chen, Tan, Wang, and Hu]{chen2018reverse}
Shuhan Chen, Xiuli Tan, Ben Wang, and Xuelong Hu.
\newblock Reverse attention for salient object detection.
\newblock In \emph{ECCV}, pages 234--250, 2018.

\bibitem[Cheng et~al.(2025)Cheng, Yin, Fu, Guo, Yang, Kautz, Wang, and Liu]{cheng2025spatialrgpt}
An-Chieh Cheng, Hongxu Yin, Yang Fu, Qiushan Guo, Ruihan Yang, Jan Kautz, Xiaolong Wang, and Sifei Liu.
\newblock Spatialrgpt: Grounded spatial reasoning in vision-language models.
\newblock \emph{NeurIPS}, 37:\penalty0 135062--135093, 2025.

\bibitem[Everingham et~al.(2010)Everingham, Van~Gool, Williams, Winn, and Zisserman]{everingham2010pascal}
Mark Everingham, Luc Van~Gool, Christopher~KI Williams, John Winn, and Andrew Zisserman.
\newblock The {P}ascal {V}isual {O}bject {C}lasses ({VOC}) challenge.
\newblock \emph{IJCV}, 88:\penalty0 303--338, 2010.

\bibitem[Gu et~al.(2022)Gu, Lin, Kuo, and Cui]{gu2022openvocabulary}
Xiuye Gu, Tsung-Yi Lin, Weicheng Kuo, and Yin Cui.
\newblock Open-vocabulary object detection via vision and language knowledge distillation.
\newblock In \emph{ICLR}, 2022.

\bibitem[Hori et~al.(2017)Hori, Hori, Lee, Zhang, Harsham, Hershey, Marks, and Sumi]{hori2017attention}
Chiori Hori, Takaaki Hori, Teng-Yok Lee, Ziming Zhang, Bret Harsham, John~R Hershey, Tim~K Marks, and Kazuhiko Sumi.
\newblock Attention-based multimodal fusion for video description.
\newblock In \emph{ICCV}, pages 4193--4202, 2017.

\bibitem[Huang et~al.(2017)Huang, Xia, Wu, Li, Wang, Song, and Kuo]{huang2017semantic}
Qin Huang, Chunyang Xia, Chihao Wu, Siyang Li, Ye Wang, Yuhang Song, and C-C~Jay Kuo.
\newblock Semantic segmentation with reverse attention.
\newblock In \emph{BMVC}, pages 18.1--18.13, 2017.

\bibitem[Hyeon-Woo et~al.(2023)Hyeon-Woo, Yu-Ji, Heo, Han, Oh, and Oh]{hyeon2023scratching}
Nam Hyeon-Woo, Kim Yu-Ji, Byeongho Heo, Dongyoon Han, Seong~Joon Oh, and Tae-Hyun Oh.
\newblock Scratching visual transformer's back with uniform attention.
\newblock In \emph{ICCV}, pages 5807--5818, 2023.

\bibitem[Jain and Wallace(2019)]{jain2019attention}
Sarthak Jain and Byron~C Wallace.
\newblock Attention is not explanation.
\newblock In \emph{Proceedings of the 2019 Conference of the North American Chapter of the Association for Computational Linguistics: Human Language Technologies, Volume 1 (Long and Short Papers)}, pages 3543--3556, 2019.

\bibitem[Kamath et~al.(2021)Kamath, Singh, LeCun, Synnaeve, Misra, and Carion]{kamath2021mdetr}
Aishwarya Kamath, Mannat Singh, Yann LeCun, Gabriel Synnaeve, Ishan Misra, and Nicolas Carion.
\newblock {MDETR}-modulated detection for end-to-end multi-modal understanding.
\newblock In \emph{ICCV}, pages 1780--1790, 2021.

\bibitem[Lee et~al.(2023)Lee, Cho, and Choi]{lee2023shallow}
Go-Eun Lee, Jungchan Cho, and Sang-II Choi.
\newblock Shallow and reverse attention network for colon polyp segmentation.
\newblock \emph{Scientific Reports}, 13\penalty0 (1):\penalty0 15243, 2023.

\bibitem[Li et~al.(2022)Li, Zhang, Zhang, Yang, Li, Zhong, Wang, Yuan, Zhang, Hwang, et~al.]{li2022grounded}
Liunian~Harold Li, Pengchuan Zhang, Haotian Zhang, Jianwei Yang, Chunyuan Li, Yiwu Zhong, Lijuan Wang, Lu Yuan, Lei Zhang, Jenq-Neng Hwang, et~al.
\newblock Grounded language-image pre-training.
\newblock In \emph{CVPR}, pages 10965--10975, 2022.

\bibitem[Li et~al.(2023)Li, Liu, Jha, and Yao]{li2023distilled}
Yun Li, Zhe Liu, Saurav Jha, and Lina Yao.
\newblock Distilled reverse attention network for open-world compositional zero-shot learning.
\newblock In \emph{ICCV}, pages 1782--1791, 2023.

\bibitem[Li et~al.(2024)Li, Yi, Uneri, Niu, and Jones]{li2024rta}
Zhikai Li, Murong Yi, Ali Uneri, Sihan Niu, and Craig Jones.
\newblock Rta-former: Reverse transformer attention for polyp segmentation.
\newblock In \emph{2024 46th Annual International Conference of the IEEE Engineering in Medicine and Biology Society (EMBC)}, pages 1--5. IEEE, 2024.

\bibitem[Liang et~al.(2021)Liang, Lin, Feng, Zhang, and Lv]{liang2021attention}
Tao Liang, Guosheng Lin, Lei Feng, Yan Zhang, and Fengmao Lv.
\newblock Attention is not enough: Mitigating the distribution discrepancy in asynchronous multimodal sequence fusion.
\newblock In \emph{ICCV}, pages 8148--8156, 2021.

\bibitem[Ren et~al.(2015)Ren, He, Girshick, and Sun]{ren2015faster}
Shaoqing Ren, Kaiming He, Ross Girshick, and Jian Sun.
\newblock Faster {R}-{CNN}: Towards real-time object detection with region proposal networks.
\newblock \emph{NeurIPS}, 28, 2015.

\bibitem[Shaw et~al.(2018)Shaw, Uszkoreit, and Vaswani]{shaw2018self}
Peter Shaw, Jakob Uszkoreit, and Ashish Vaswani.
\newblock Self-attention with relative position representations.
\newblock In \emph{Proceedings of the 2018 Conference of the North American Chapter of the Association for Computational Linguistics: Human Language Technologies, Volume 2 (Short Papers)}, pages 464--468, 2018.

\bibitem[Shen et~al.(2024)Shen, Zhao, Zhu, and Yin]{shen2024groundvlp}
Haozhan Shen, Tiancheng Zhao, Mingwei Zhu, and Jianwei Yin.
\newblock Ground{VLP}: Harnessing zero-shot visual grounding from vision-language pre-training and open-vocabulary object detection.
\newblock In \emph{AAAI}, pages 4766--4775, 2024.

\bibitem[Sinhamahapatra et~al.(2025)Sinhamahapatra, Bose, Roscher, and G{\"u}nnemann]{sinhamahapatra2025zero}
Poulami Sinhamahapatra, Shirsha Bose, Karsten Roscher, and Stephan G{\"u}nnemann.
\newblock Zero-shot open-vocabulary {OOD} object detection and grounding using vision language models.
\newblock In \emph{Northern Lights Deep Learning Conference}, pages 230--238. PMLR, 2025.

\bibitem[Sun et~al.(2019)Sun, Kim, Lee, Kim, and Ko]{sun2019reverse}
Jee-Young Sun, Seung-Wook Kim, Sang-Won Lee, Ye-Won Kim, and Sung-Jea Ko.
\newblock Reverse and boundary attention network for road segmentation.
\newblock In \emph{ICCV}, pages 0--0, 2019.

\bibitem[Venhoff et~al.(2025)Venhoff, Khakzar, Joseph, Torr, and Nanda]{venhoff2025visual}
Constantin Venhoff, Ashkan Khakzar, Sonia Joseph, Philip Torr, and Neel Nanda.
\newblock How visual representations map to language feature space in multimodal llms.
\newblock \emph{arXiv preprint arXiv:2506.11976}, 2025.

\bibitem[Wang et~al.(2024{\natexlab{a}})Wang, Feng, Li, Cheng, Lyu, Liu, Chen, and Zhao]{wang2024ov}
Chunlei Wang, Wenquan Feng, Xiangtai Li, Guangliang Cheng, Shuchang Lyu, Binghao Liu, Lijiang Chen, and Qi Zhao.
\newblock {OV}-{VG}: A benchmark for open-vocabulary visual grounding.
\newblock \emph{Neurocomputing}, 591:\penalty0 127738, 2024{\natexlab{a}}.

\bibitem[Wang et~al.(2024{\natexlab{b}})Wang, Xie, Yang, and Song]{wang2024ra}
Zhenyuan Wang, Xuemei Xie, Jianxiu Yang, and Xiaodan Song.
\newblock Ra-net: reverse attention for generalizing residual learning.
\newblock \emph{Scientific Reports}, 14\penalty0 (1):\penalty0 12771, 2024{\natexlab{b}}.

\bibitem[Wiegreffe and Pinter(2019)]{wiegreffe2019attention}
Sarah Wiegreffe and Yuval Pinter.
\newblock Attention is not not explanation.
\newblock In \emph{2019 Conference on Empirical Methods in Natural Language Processing and 9th International Joint Conference on Natural Language Processing, EMNLP-IJCNLP 2019}, pages 11--20. Association for Computational Linguistics, 2019.

\bibitem[Wu et~al.(2025)Wu, Chen, Luo, Zhang, Gao, He, Wang, Lin, and Qiu]{wu2025valley2}
Ziheng Wu, Zhenghao Chen, Ruipu Luo, Can Zhang, Yuan Gao, Zhentao He, Xian Wang, Haoran Lin, and Minghui Qiu.
\newblock Valley2: Exploring multimodal models with scalable vision-language design.
\newblock \emph{arXiv preprint arXiv:2501.05901}, 2025.

\bibitem[Xie et~al.(2019)Xie, Liu, Li, Cheng, Zuo, Liu, Wen, and Ding]{xie2019image}
Chaohao Xie, Shaohui Liu, Chao Li, Ming-Ming Cheng, Wangmeng Zuo, Xiao Liu, Shilei Wen, and Errui Ding.
\newblock Image inpainting with learnable bidirectional attention maps.
\newblock In \emph{ICCV}, pages 8858--8867, 2019.

\bibitem[Yang et~al.(2025)Yang, Ma, Wang, Yin, Rong, Rao, and Zhang]{yang2025wethink}
Jie Yang, Feipeng Ma, Zitian Wang, Dacheng Yin, Kang Rong, Fengyun Rao, and Ruimao Zhang.
\newblock We{T}hink: Toward general-purpose vision-language reasoning via reinforcement learning.
\newblock \emph{arXiv preprint arXiv:2506.07905}, 2025.

\bibitem[Yu et~al.(2025)Yu, Zhang, Li, Xu, Yao, Chen, Lu, Cui, Dang, He, et~al.]{yu2025rlaif}
Tianyu Yu, Haoye Zhang, Qiming Li, Qixin Xu, Yuan Yao, Da Chen, Xiaoman Lu, Ganqu Cui, Yunkai Dang, Taiwen He, et~al.
\newblock {RLAIF-V}: Open-source {AI} feedback leads to super {GPT-4V} trustworthiness.
\newblock In \emph{CVPR}, pages 19985--19995, 2025.

\bibitem[Zareian et~al.(2021)Zareian, Rosa, Hu, and Chang]{zareian2021open}
Alireza Zareian, Kevin~Dela Rosa, Derek~Hao Hu, and Shih-Fu Chang.
\newblock Open-vocabulary object detection using captions.
\newblock In \emph{CVPR}, pages 14393--14402, 2021.

\bibitem[Zhou et~al.(2025)Zhou, Zhao, Chen, Wang, and Metaxas]{zhou2025led}
Yang Zhou, Shiyu Zhao, Yuxiao Chen, Zhenting Wang, and Dimitris~N Metaxas.
\newblock {LED}: {LLM} enhanced open-vocabulary object detection without human curated data generation.
\newblock \emph{arXiv preprint arXiv:2503.13794}, 2025.

\end{thebibliography}
}

\clearpage
\setcounter{page}{1}
\maketitlesupplementary
\renewcommand{\thefigure}{S\arabic{figure}}
\renewcommand{\thesection}{S\arabic{section}}



\section{FitAP}
\label{section:fitap}
Based on standard definitions, the FitAP, similar to the mean average precision in object detection, can be defined as
\begin{equation*}
    \text{FitAP} = \frac{1}{10}\sum_{i=1}^{10} \text{AP}(\Theta_i), 
\end{equation*}
wherein $\mathbf{\Theta}=\left\{\Theta_i=0.5+(i-1)0.05\; |\; i=1,2,\ldots,10\right\}$.

In the absence of a confidence score from the parsed VLM detection results, we propose to use the product of the normalized box area of detection and the IoU, $A_{\mathrm{box}}\times \mathrm{IoU}$, for the quality ranking of detection against the ground truth data. 

We establish this approach by first showing the correlations of the area of the ground truth boxes with $A_{\mathrm{box}}$ and $A_{\mathrm{box}}\times \mathrm{IoU}$. Then, we visualize samples of the precision-recall curves generated by this approach, pointing out how its features resemble those generated by vision-only models. From these, we confirm that $A_{\mathrm{box}}\times \mathrm{IoU}$ is a reliable substitute for traditional confidence scores in generating precision-recall curves and calculating AP. 

The critical step is to ensure that the metric used reliably correlates with the probability of detection being \emph{true positive}, which is crucial to accurately calculate the AP and understand the performance of the model.

\subsection{Area correlations}
\begin{figure*}[htbp]
  \centering
  \includegraphics[width=\textwidth]{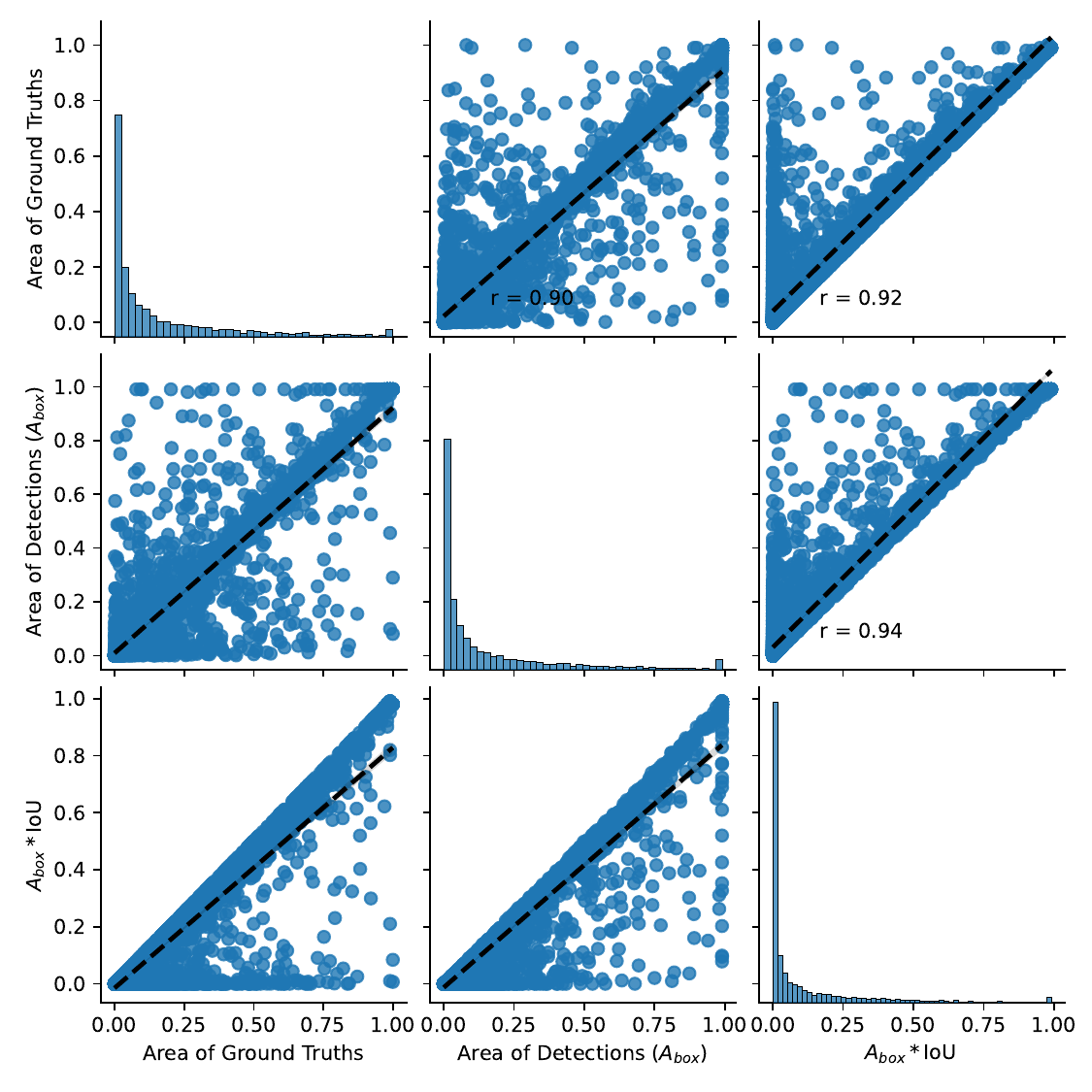}
  \caption{Correlations of ground-truth box area with $A_{\mathrm{box}}$ and $A_{\mathrm{box}}\times\mathrm{IoU}$ from the output of VLM queried with $p_5$. The plots include results from all object categories taken at IoU threshold, $\Theta = 0.50$. Diagonals are the univariate histograms of ground-truth area, $A_{\mathrm{box}}$, and $A_{\mathrm{box}}\times\mathrm{IoU}$. The dashed lines represent linear regression fits with the Pearson correlation coefficient, $r$, shown only for the upper triangular plots.}
  \label{fig:correlations}
\end{figure*}

Here, we offer an empirical basis of $A_{\mathrm{box}}\times\mathrm{IoU}$ for the quality ranking of the detection boxes by VLM from the following:
\begin{enumerate}
    \item[1.] The tendency of VLM's predictions to maintain proportional sizing with the actual object in the image, and
    \item[2.] influence of the actual object's size on the detection accuracy.
\end{enumerate}
Thus, we examine the correlation between the area of ground truths (normalized to image size) and those of detection $A_{\mathrm{box}}$. Our results (Figure~\ref{fig:correlations}) confirm that this correlation is strong (Pearson $r = 0.90$), indicating that VLM tends to generate detections with areas similar to the actual objects, affirming VLM's sizing accuracy, which is an essential aspect of \textbf{objectness}. The size accuracy implies that VLM recognizes and localizes the actual object in the image.

We further establish the correlation between the ground truth box areas and the proposed metric $A_{\mathrm{box}}\times\mathrm{IoU}$. The results (Figure~\ref{fig:correlations}) also confirm that this correlation is strong (Pearson $r=0.92$), suggesting that larger, more well-fitting boxes are more common when the model correctly detects objects. In fact, this metric captures both the size and the quality of fit of the detections.

\subsection{Sample precision-recall curves}
\begin{figure*}[htbp]
  \centering
  \begin{subfigure}{0.27\textwidth}
    \includegraphics[width=\textwidth]{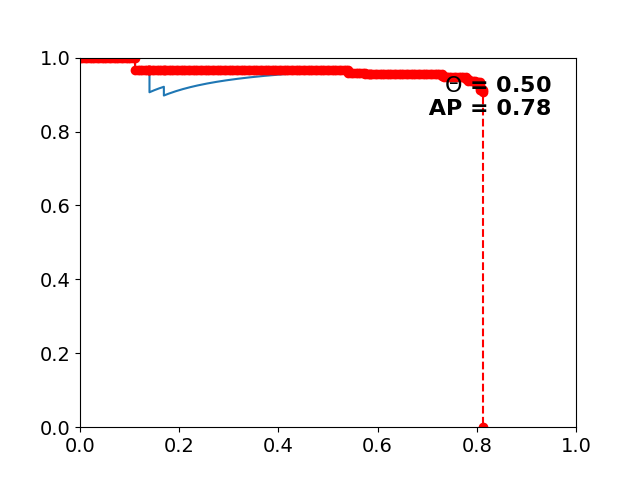}
    \caption{$\Theta = 0.50$, $\mathrm{AP} = 0.78$}
  \end{subfigure}
  \begin{subfigure}{0.27\textwidth}
    \includegraphics[width=\textwidth]{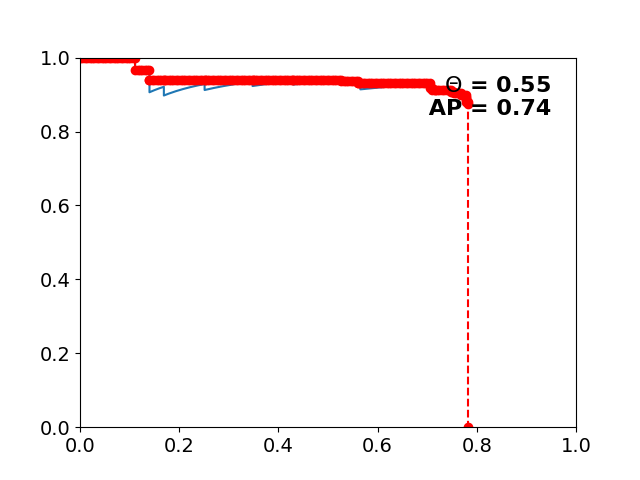}
    \caption{$\Theta = 0.55$, $\mathrm{AP} = 0.74$}
  \end{subfigure}
  \begin{subfigure}{0.27\textwidth}
    \includegraphics[width=\textwidth]{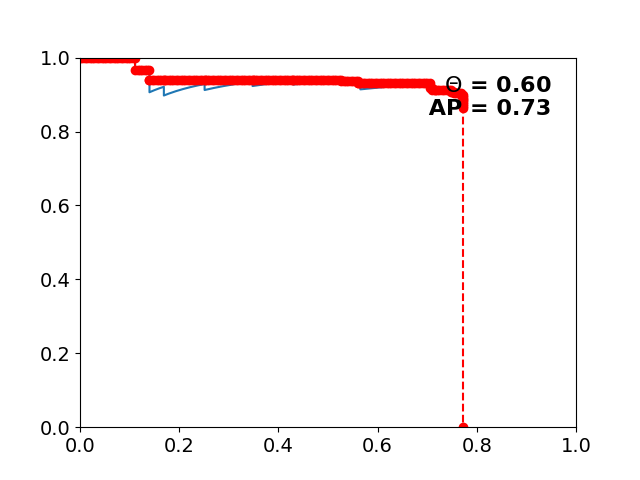}
    \caption{$\Theta = 0.60$, $\mathrm{AP} = 0.73$}
  \end{subfigure}
  \begin{subfigure}{0.27\textwidth}
    \includegraphics[width=\textwidth]{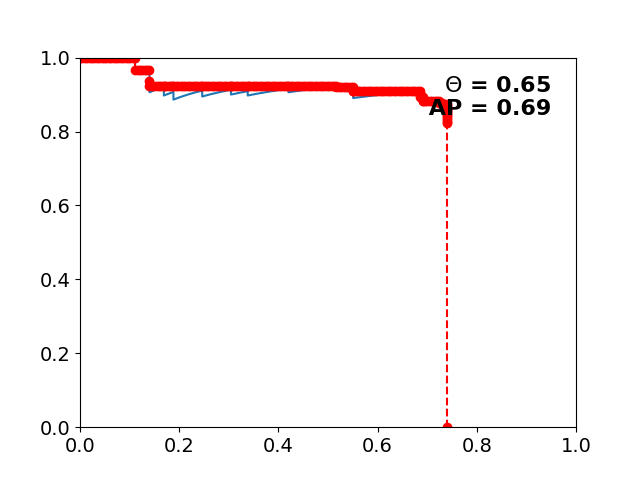}
    \caption{$\Theta = 0.65$, $\mathrm{AP} = 0.69$}
  \end{subfigure}
  \begin{subfigure}{0.27\textwidth}
    \includegraphics[width=\textwidth]{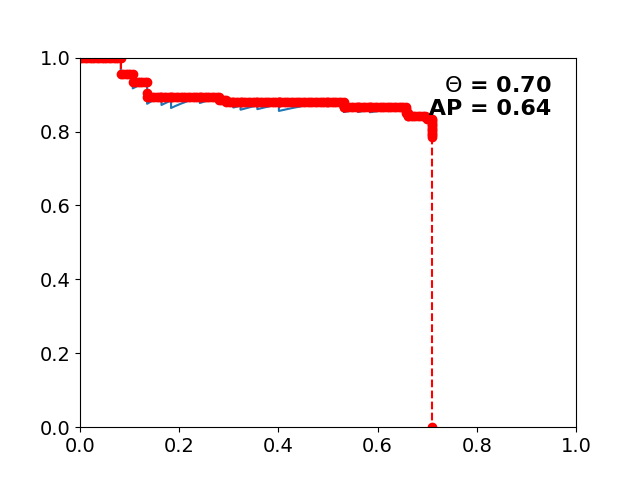}
    \caption{$\Theta = 0.70$, $\mathrm{AP} = 0.64$}
  \end{subfigure}
  \begin{subfigure}{0.27\textwidth}
    \includegraphics[width=\textwidth]{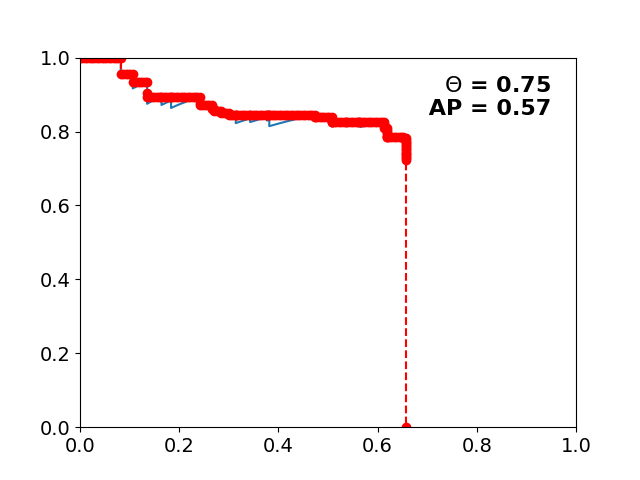}
    \caption{$\Theta = 0.75$, $\mathrm{AP} = 0.57$}
  \end{subfigure}
  \begin{subfigure}{0.27\textwidth}
    \includegraphics[width=\textwidth]{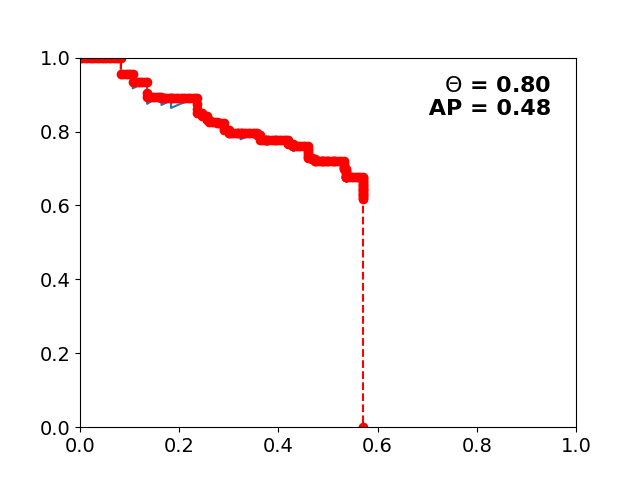}
    \caption{$\Theta = 0.80$, $\mathrm{AP} = 0.48$}
  \end{subfigure}
  \begin{subfigure}{0.27\textwidth}
    \includegraphics[width=\textwidth]{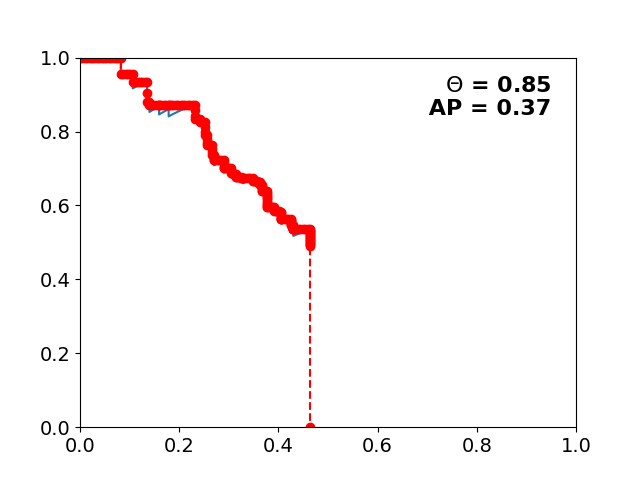}
    \caption{$\Theta = 0.85$, $\mathrm{AP} = 0.37$}
  \end{subfigure}
  \begin{subfigure}{0.27\textwidth}
    \includegraphics[width=\textwidth]{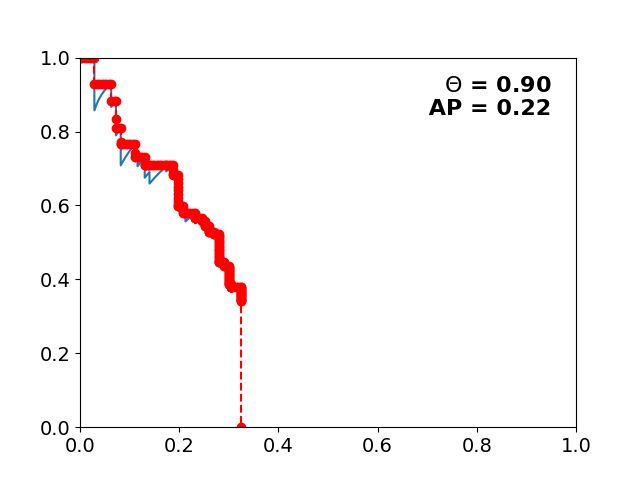}
    \caption{$\Theta = 0.90$, $\mathrm{AP} = 0.22$}
  \end{subfigure}
  \begin{subfigure}{0.27\textwidth}
    \includegraphics[width=\textwidth]{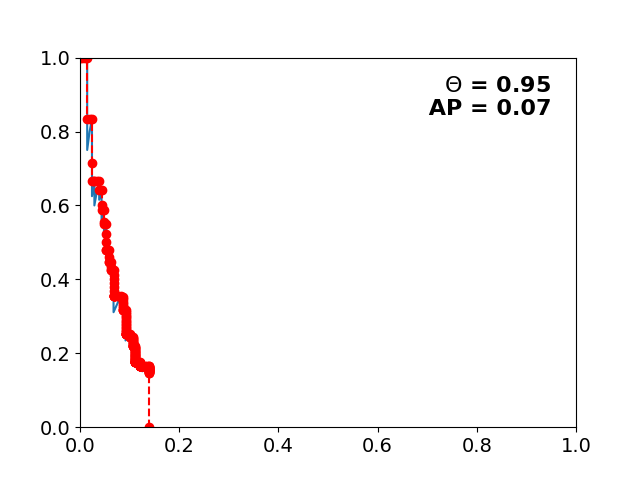}
    \caption{$\Theta = 0.95$, $\mathrm{AP} = 0.07$}
  \end{subfigure}
  \caption{Precision-recall curves for the category \texttt{giraffe} at different IoU thresholds $\Theta$ and the corresponding average precision, AP (area under the curve). Solid (blue) curves from actual data; dashed lines-points (red) represent envelopes from which FitAP is calculated as the average of AP for $\Theta\in[0.50:0.05:0.95]$.}
  \label{fig:prcurves}
\end{figure*}

The choice of a metric to replace confidence scores should ideally reflect the confidence in detections being true positives. By showing samples of the generated precision-recall curves, we empirically demonstrate that $A_{\mathrm{box}}\times\mathrm{IoU}$ correlates with actual detection performance and does not introduce bias or misrepresentation in model evaluation.

The first noticeable feature is the general decreasing trends shown in Figure~\ref{fig:prcurves}. This trend expresses the expected trade-off between precision and recall. Attempting to fit tighter (more precise) boxes increases the tendency to miss actual objects (less recall). However, aiming for better recall comes at the expense of looser detections.

Another peculiar feature displayed in Figure~\ref{fig:prcurves}(h), (i) is the zigzag pattern of the empirical curve. The zigzag is an artifact of deriving floating-point ratios, i.e. precision and recall, from counting. As we aim for better recall, more detections are necessary at the expense of some of these being false positives, which explains the abrupt vertical drops. Gradual recovery is attributed to the acquisition of true positives and the improvement in recall. Then, another peak is encountered, at which point the next drop-off starts. However, succeeding peaks are, nevertheless, getting lower such that the envelope maintains the downward trajectory of the curve.

Finally, notice how the AP correspondingly decreases as the IoU threshold $\Theta$ increases. This tradeoff is evident from the curve's displacement toward the plot's bottom-left corner. This displacement effectively reduces the area under the curve, hence reducing FitAP. Higher $\Theta$ expresses a stricter criterion to detect true positives, resulting in fewer correct detections.

The precision-recall curves for other categories display the same characteristics. Therefore, we have shown how well $A_{\mathrm{box}}\times\mathrm{IoU}$ performs in predicting true positives, making it applicable for evaluating the object detection capability of VLMs.

\begin{figure*}[htbp]
  \centering
  \begin{minipage}{0.39\linewidth}
    \centering
    \includegraphics[width=\linewidth]{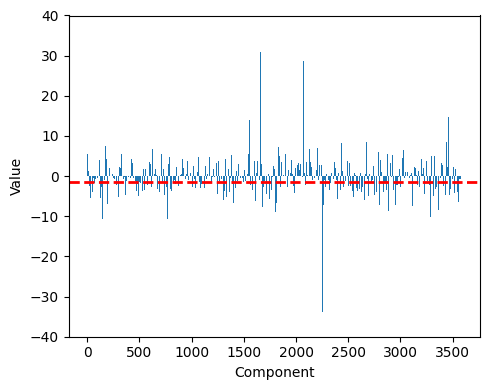}
    \caption*{(a) Pre-RCA hidden state $z_i$}
  \end{minipage}
  \hspace{0.02\linewidth}
  \begin{minipage}{0.16\linewidth}
    \centering
    \vspace{-0.5\linewidth}
${\Huge\implies\boxed{\widetilde{A}}\implies}$
  \end{minipage}
  \hspace{0.02\linewidth}
  \begin{minipage}{0.39\linewidth}
    \centering
    \includegraphics[width=\linewidth]{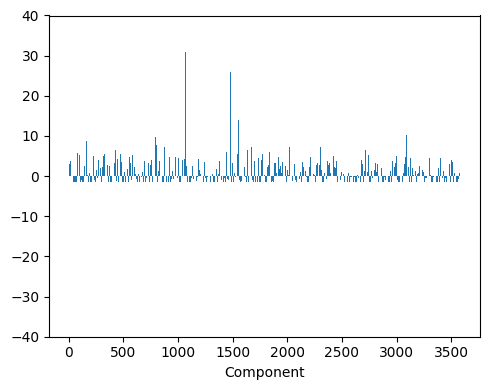}
    \caption*{(b) Post-RCA hidden state $\widetilde{z_i}$}
  \end{minipage}
  \caption{Flooring the subthreshold contributions of a hidden state $z_i$ implicitly implies $\widetilde{A}$ leading to $\widetilde{z_i}$. The red dashed horizontal line in (a) corresponds to $\vartheta=-1.5$. In this example, the embedding size is $d=3584$}
  \label{fig:flooring-example}
\end{figure*}

\section{Indicators of the RCA-driven improvement}
\label{section:RCA-correlations}
Here we develop a formal mathematical argument discussion that shows how Condition~(\ref{ineq:formal-flooring-inequality}) from the paper establishes a negative relationship between the number of subthreshold contributions to the hidden state and the scaler $m$, which is the mean cross-head maximum of attention weights. We also prove that this inverse relationship is valid regardless of whether RCA uses inverse-distance or Gaussian peak reweighting, as defined in Section~\ref{sec:rca}. For the preliminaries, let:
\begin{itemize}
    \item $\alpha_{ij}\in[0,1]$: the base attention weights from token $i$ to $j$
    \item $A^{(h)}\in \mathbb{R}^{n\times n}$: attention map from head $h$, for $h=1,\ldots,H$
    \item $A_{ij}^{\text{max}}:=\max_h A_{ij}^{(h)}$
    \item $m:=\frac{1}{n}\sum_{j=1}^n\max_iA_{ij}^{\text{max}}$: mean column-maximum of $A^{\text{max}}$
\end{itemize}
The value of $m$ quantifies the \textbf{global sharpness} in attention across all heads. 

We want to show that:
\begin{equation*}
    m\uparrow \implies \sum_{j\in \mathcal{J_{\downarrow}}}\widetilde{\alpha_{ij}}\,\downarrow\implies  \widetilde{z_i}(d)\uparrow
\end{equation*}
implying fewer subthreshold components, and thus, condition~(\ref{ineq:formal-flooring-inequality}) implies a \textbf{negative relationship} between the number of subthreshold components and $m$. 

From the paper:
\begin{equation*}
    \widetilde{z_i}(d)\geq \vartheta + \left(v^{-}-\vartheta\right)\sum_{j\in\mathcal{J}_{\downarrow}}\widetilde{\alpha_{ij}},
\end{equation*}
where:
\begin{itemize}
    \item $\vartheta$: threshold value (floor)
    \item $v^{-}<\vartheta$: minimal value component of subthreshold tokens
    \item $\widetilde{\alpha_{ij}}$: RCA-transformed attention weights (depends on $m$)
\end{itemize}
Thus, minimizing $\sum_{j\in\mathcal{J}_{\downarrow}}$ tightens the bound so that $\widetilde{z_i}(d)$ is closer to or above $\vartheta$. For the key strategy, we show:
\begin{enumerate}
    \item $m\uparrow\implies \widetilde{\alpha_{ij}}$ assigns less weight to $j\in\mathcal{J}_{\downarrow}$.
    \item This assertion holds for both RCA schemes:
    \begin{itemize}
    \item Inverse-distance from $m$
    \item Gausian peaking around $m$
    \end{itemize}
\end{enumerate}
Therefore, the penalty term in condition~(\ref{ineq:formal-flooring-inequality}) shrinks with increasing $m$, which increases $\widetilde{z_i}(d)$, decreasing the subthreshold count.

In the first case of inverse-distance reweighting:
\begin{equation*}
    \alpha'_{ij} = \frac{1}{1+\gamma|\alpha_{ij}-m|},
\end{equation*}
which peaks at $\alpha_{ij}=m$ and decreases as $\alpha_{ij}$ deviates from $m$. Suppose $m$ increases. Then for fixed $\alpha_{ij}$, the distance $|\alpha_{ij}-m|$ increases unless $\alpha_{ij}$ tracks $m$. Thus, for subthreshold contributors $j\in\mathcal{J}_{\downarrow}$, which typically have $\alpha_{ij}<m$ and $v_j(d)<\vartheta$, we get:
\begin{equation*}
    \alpha'_{ij}(m)\downarrow\implies\widetilde{\alpha_{ij}}\downarrow\implies\sum_{j\in\mathcal{J}_{\downarrow}}\widetilde{\alpha_{ij}}\downarrow\implies\widetilde{z_i}(d)\uparrow,
\end{equation*}
implying that the subthreshold count decreases.

In the second case of Gaussian peak reweighting:
\begin{equation*}
    \alpha'_{ij} = \exp\left[-\gamma\left(\alpha_{ij}-m\right)^2\right],
\end{equation*}
which symmetrically peaks at $\alpha_{ij}=m$ and rapidly decays as $\alpha_{ij}$ moves away from $m$. Suppose $m$ increases. For fixed $\alpha_{ij}$, again $|\alpha_{ij}-m|$ increases and so $\alpha'_{ij}$ decreases and penalizes values further away from $m$. Thus, subthreshold tokens $j\in\mathcal{J}_{\downarrow}$ with mid- or low $\alpha_{ij}$, get decreasing attention as $m$ increases. So again,
\begin{equation*}
    \sum_{j\in\mathcal{J}_{\downarrow}}\widetilde{\alpha_{ij}}\downarrow\implies\widetilde{z_i}(d)\uparrow\implies\text{ subthreshold count}\downarrow.
\end{equation*}

From these arguments, we have shown that under both RCA reweighting strategies (inverse-distance and Gaussian peaking), as $m\uparrow$, subthreshold tokens $j\in\mathcal{J}_{\downarrow}$ receive less attention mass so $\sum_{j\in\mathcal{J}\downarrow}\widetilde{\alpha_{ij}}\downarrow$, which increases the lower bound of Condition~(\ref{ineq:formal-flooring-inequality}). Thus, decreasing the number of components $\widetilde{z_i}$ that fall below $\vartheta$, as visualized in Fig.~\ref{fig:flooring-example}
\begin{equation*}
    \boxed{\frac{d|S|(\widetilde{z_i})}{dm} < 0\; \text{  as implied by Condition~(\ref{ineq:formal-flooring-inequality})}}
\end{equation*}
where $|S|$ is the number of subthreshold contributors. This conclusion establishes that condition~(\ref{ineq:formal-flooring-inequality}) supports a negative relationship between the subthreshold count and the attention sharpness measure $m$, regardless of RCA variant used. 


\section{Online Repository}
The codes and data sets used by this study are accessible from \url{https://github.com/earl-juanico/rca}


\end{document}